\documentclass{article}
\usepackage{ijcai25}

\usepackage{times}
\usepackage{soul}
\usepackage{url}
\usepackage[hidelinks]{hyperref}
\usepackage[utf8]{inputenc}
\usepackage[small]{caption}
\usepackage{graphicx}
\usepackage{amsmath}
\usepackage{amsthm}
\usepackage{booktabs}
\usepackage{algorithm}
\usepackage{algorithmic}
\usepackage[switch]{lineno}
\usepackage[capitalise,noabbrev]{cleveref} 
\crefname{appsec}{Appendix}{Appendices} 
\usepackage{framed}
\usepackage{subcaption}
\usepackage{xcolor}

\usepackage{enumitem}
\usepackage{comment}

\title{A Picture is Worth a Thousand Prompts? Efficacy of Iterative Human-Driven Prompt Refinement in Image Regeneration Tasks}

\author{
Khoi Trinh\textsuperscript{\rm 1},
Scott Seidenberger\textsuperscript{\rm 1},
Raveen Wijewickrama\textsuperscript{\rm 2},
Murtuza Jadliwala\textsuperscript{\rm 2},
Anindya Maiti\textsuperscript{\rm 1} \\
\textsuperscript{\rm 1} \textnormal{University of Oklahoma} \\
\textsuperscript{\rm 2} \textnormal{University of Texas at San Antonio}\\
\textnormal{\textit{khoitrinh@ou.edu, seidenberger@ou.edu, raveen.wijewickrama@utsa.edu, murtuza.jadliwala@utsa.edu, am@ou.edu}}
}

\begin{document}

\maketitle

\begin{abstract}

With AI-generated content becoming ubiquitous across the web, social media, and other digital platforms, it is vital to examine how such content are inspired and generated.
The creation of AI-generated images often involves refining the input prompt iteratively to achieve desired visual outcomes. This study focuses on the relatively underexplored concept of image \emph{regeneration} using AI, in which a human operator attempts to closely recreate a specific target image by iteratively refining their prompt. Image regeneration is distinct from normal image generation, which lacks any predefined visual reference. A separate challenge lies in determining whether existing image similarity metrics (ISMs) can provide reliable, objective feedback in iterative workflows, given that we do not fully understand if subjective human judgments of similarity align with these metrics. Consequently, we must first validate their alignment with human perception before assessing their potential as a feedback mechanism in the iterative prompt refinement process. To address these research gaps, we present a structured user study evaluating how iterative prompt refinement affects the similarity of regenerated images relative to their targets, while also examining whether ISMs capture the same improvements perceived by human observers. Our findings suggest that incremental prompt adjustments substantially improve alignment, verified through both subjective evaluations and quantitative measures—underscoring the broader potential of iterative workflows to enhance generative AI content creation across various application domains.

\end{abstract}

\section{Introduction}
\label{sec:intro}

The rise of AI-generated content on online platforms has made it crucial to investigate how this type of content is created, specifically through the iterative processes of image generation and regeneration. While prior work has explored AI-led iterative refinement, this paper highlights the human's leading role in refining prompts and improving outcomes through their own judgment and control. The field of generative artificial intelligence (GenAI) has recently seen significant advancements, particularly in the development of text-to-image (\texttt{txt2img}) models. These models provide an easy and fast process for creating high-quality artwork. Among the notable \texttt{txt2img} models contributing to this trend are Midjourney \cite{midjourney}, DALL-E 3 \cite{betker2023improving}, and Stable Diffusion 3 \cite{esserstable}. While these models enable a more accessible way to generate high-quality and visually appealing images, creating artwork with this method (specifically, \textit{image generation}) is usually iterative. A user starts with a concept, formulates a prompt for the \texttt{txt2img} model, and uses the prompt as input to the model to obtain the desirable image. If the obtained image is unsatisfactory, the user repeatedly refines the prompt until the desired result is achieved or the user abandons the task.

\textit{Image regeneration} through prompt refinement refers to a task where a user iteratively edits their prompt with the goal of recreating a visual based on some target image or visual style. This iterative process illustrates human-AI interaction techniques, where the user and AI collaborate to achieve optimal outputs. By iteratively refining prompts, users actively guide the creative process, demonstrating the potential of human-AI collaboration to bridge gaps between technical capabilities and artistic intent.
This concept of image regeneration through iterative prompt refinement has numerous practical applications, such as bypassing prompt marketplaces, educating novice users, restoration of lost or damaged art pieces
\cite{trinh2024promptly,tang2024exploring,oppenlaender2024prompting,kulkarni2023word,liang2023iterative}. Despite the potential applications, limited research exists on how humans can improve image quality through iterative prompt refinement.

Iterative refinement processes have been utilized by humans in a wide domain of tasks such as writing, programming, and design. Flower \cite{flower1981cognitive} provides a model on how writers plan, create, and revise their work iteratively. Madaan \cite{madaan2024self} show that iterative refinement is effective in significantly improving outputs in text and code generation tasks through self-feedback mechanisms. M{\o}ller and Aiello \cite{moller2024prompt} show that stepwise prompt refinement can show improvement in text summarization tasks. Du \cite{du2022read} provides the R3 framework that has demonstrated the effectiveness of iterative revision in producing high-quality textual outputs by incorporating user feedback at each stage of the revision. For \texttt{txt2img} generation, automatic prompt optimization systems have demonstrated substantial improvements in image quality by refining prompts systematically \cite{manas2024improving}. Building on these findings, this paper explores how iterative refinement impacts human-guided prompt optimization in image regeneration tasks, particularly in aligning generated visuals with target images.

In image regeneration and comparison, image similarity metrics (ISMs) such as Perceptual Similarity \cite{zhang2018perceptual}, CLIP scores \cite{clip,wang2023exploring}, and ImageHash \cite{imagehash} can provide objective feedback on the likeness between two images \cite{saharia2022image,zhang2018perceptual}. However, these ISMs' alignment with humans' subjective judgment remains untested. 
Since humans are the ultimate decision-makers in creative workflows, it is critical to ensure that ISMs align with subjective human evaluations. Humans make the final call on whether AI-generated outputs meet their intended purpose, making human agreement essential to validate the reliability and practical applicability of ISMs.
To address this, we first seek to understand the alignment of ISMs with human perception by comparing the objective rankings provided by these metrics to users' subjective rankings of the similarity between generated images and target images. This evaluation is vital for determining the feasibility of using ISMs as reliable feedback tools in iterative prompt refinement workflows.

Previous work by Trinh \cite{trinh2024promptly} has studied how humans' inference may compare to machine inference (i.e CLIP interrogator), and shows that while humans are able to infer prompts and generate similar images, their efforts were not as effective as using the original target prompt. However, this previous research was limited to single-shot inference, where participants had only one attempt to generate a prompt. Furthermore, image similarity metrics (ISMs) were employed to define a threshold for successful inference, considering the task complete if the generated image achieved an ISM score above the threshold. In contrast to this prior work, our study seeks to examine how participants improve image regeneration performance when allowed multiple iterations to refine prompts. Instead of using ISMs solely to determine task completion, we leverage these metrics to quantify the iterative improvements in image regeneration. This allows us to investigate the effectiveness of iterative refinement as an approach to improving human-guided AI image generation.

From these research gaps and motivations, we conducted an experiment with human subject participants to assess their improvement in an image regeneration task
This study also serves to provide additional insights on the alignment of different ISMs with subjective human assessment. We seek to address the following specific research questions (RQ):

\begin{itemize}[noitemsep, topsep=0pt, left=0pt]
    \item \textbf{RQ1:} Do humans agree that ISMs are reliable for image similarity evaluation? %
    \item \textbf{RQ2:} Does iterative prompt refinement improve both subjective and objective measures of humans in image regeneration tasks?
\end{itemize}

\noindent Our contributions in this paper are as follows:
\begin{enumerate}[noitemsep, topsep=0pt, left=0pt]
    \item \textbf{Survey Deployment:} We conducted an in-person experiment with 20 human subject participants from our host institution.
    \item \textbf{Survey Data:} Each participant conducted iterative prompt refinements for 10 target images over 10 iterations each, generating a total of 2000 prompts.
    \item \textbf{Data Evaluation:} We utilized a comprehensive set of metrics in our evaluation, including Intraclass Correlation Coefficient to gauge the alignment of ISMs to human assessment. A mixed-effects model was used to analyze ISMs in quantifying the performance improvement of the iterative prompt refinement task. Additionally, we assess the iteration at which the highest user-ranked images were generated, as an additional metric to quantify improvement in iterative prompt refinement.
\end{enumerate}

\section{Background \& Related Work}
\label{sec:related}

\begin{figure}[b]
\centering
\includegraphics[width=0.99\linewidth]{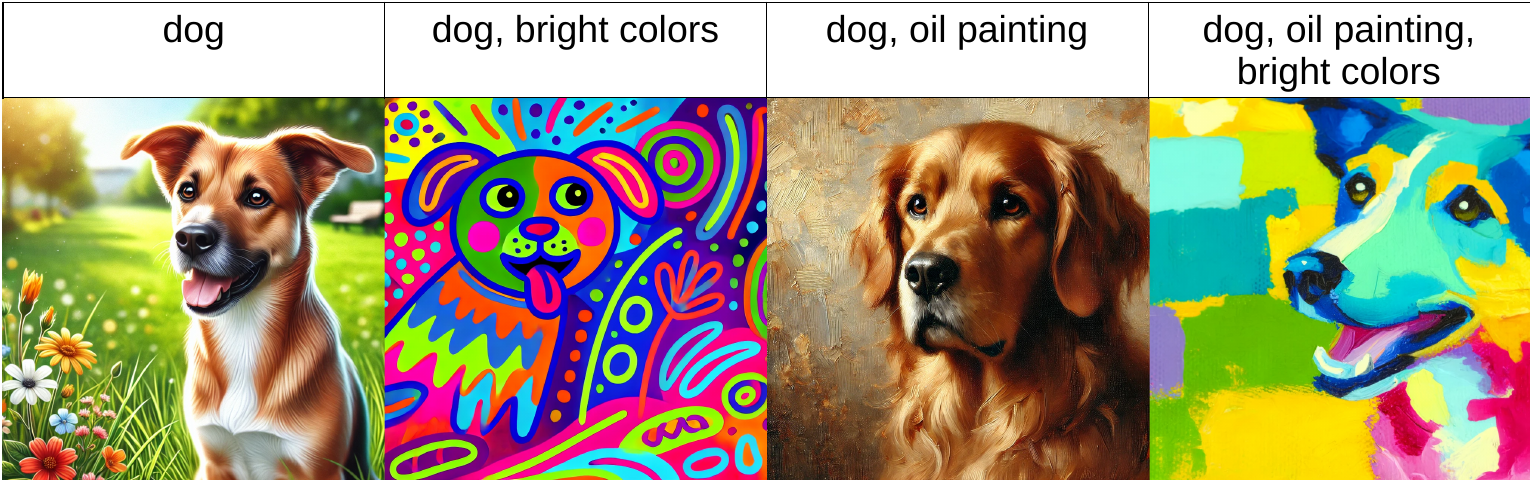}
\caption{Image generations using DALL-E 3 with prompts containing the same subject (dog) and different combinations of two modifiers (oil painting and bright colors).}
\label{fig:sub-mod-example}
\end{figure}

\begin{figure*}[ht]
\centering
\includegraphics[width=0.9\linewidth]{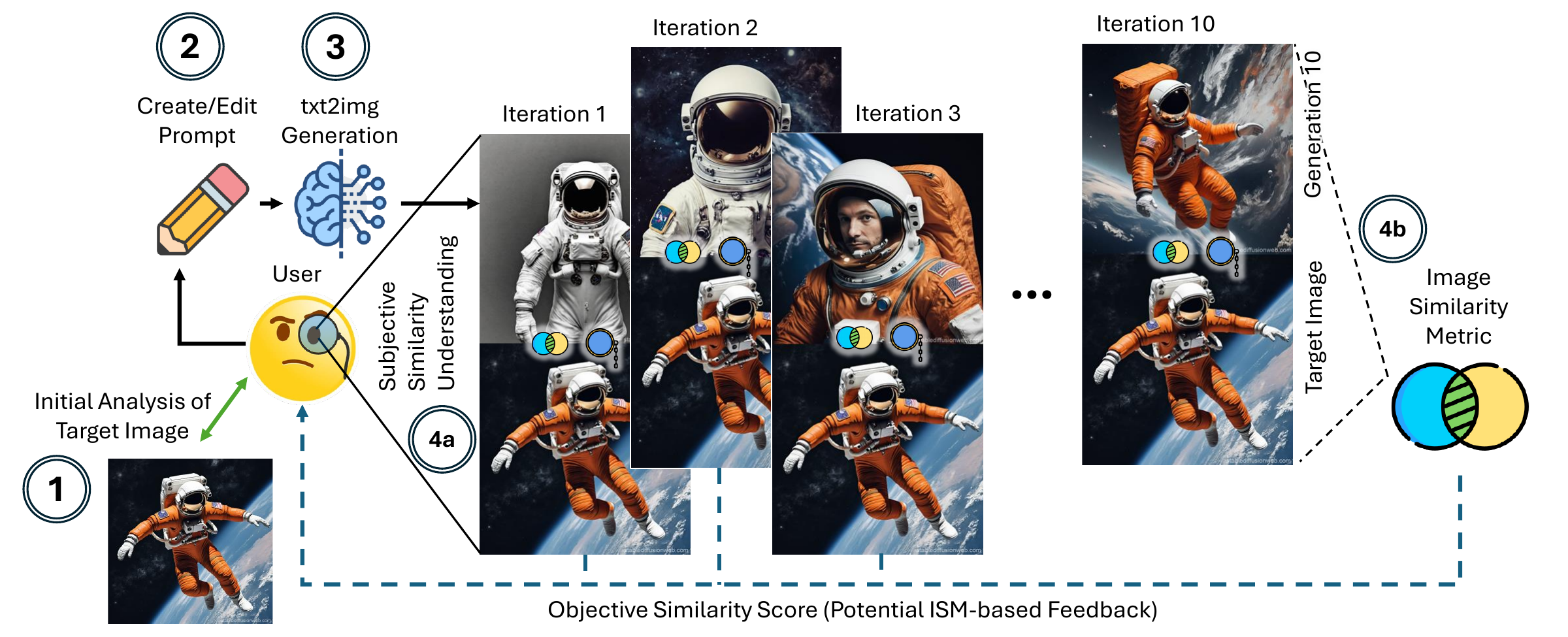} %
\caption{Summary of the iterative prompt refinement process for image regeneration task.} 
\label{fig:study-sum}
\end{figure*}

\subsection{AI Image Generation} %

Modern \texttt{txt2img} systems often pair a language encoder (e.g., a Transformer-based or large language model) with a generative model, such as a diffusion model, a Generative Adversarial Network (GAN), or a Variational Autoencoder (VAE) \cite{jia2024human}. In diffusion-based approaches (e.g., Stable Diffusion \cite{rombach2022high}), the model iteratively denoises a latent representation conditioned on the text embedding, ultimately decoding it into pixel space. GAN-based frameworks, by contrast, feature a generator trained adversarially against a discriminator \cite{goodfellow2020generative}, driving the production of increasingly realistic images. VAEs encode input data into a latent space and then decode the latent vectors back into images, facilitating synthesis from sampled noise \cite{jia2024human}.

Despite these technical advances, small prompt modifications can produce markedly different outputs \cite{trinh2024promptly}. \Cref{fig:sub-mod-example} demonstrates how substituting or reordering modifiers (e.g., \textit{bright colors, oil painting}) in a prompt describing a \textit{dog} can lead to substantial stylistic variations. This sensitivity underscores the need for systematic techniques, such as iterative prompt refinement, to achieve precise user objectives.

\subsection{Iterative Prompt Refinement}

While image generation often involves exploratory creativity, image regeneration introduces a more structured approach: recreating a specific target visual through iterative prompt refinement. This task requires users to refine their prompt iteratively, guided by feedback, either in the form of subjective assessment by the user or objective values such as ISMs, to achieve closer alignment with the target visual.

As illustrated in \cref{fig:study-sum}, the process begins with the user analyzing the target image (Step 1) to identify key visual elements and features that need to be replicated. Based on this analysis, the user creates a text prompt (Step 2) and submits it to the txt2img generation model (Step 3), which produces an initial image. Following each generation, the process involves a subjective similarity assessment (Step 4a), where the user evaluates how closely the generated image matches the target image and refines the prompt accordingly. Additionally, a potential objective similarity assessment (Step 4b) uses image ISMs to provide a quantified feedback score. Steps 2, 3, 4a, and 4b are repeated iteratively, while the user edits the prompt in each iteration, and by the 10th iteration, the goal is to generate an image that closely aligns with the target image, reflecting improvements guided by human judgment and additional ISM feedback.

Image regeneration through prompt creation by humans can have several use cases. Trinh \cite{trinh2024promptly} presented human prompt inference as a way to bypass prompt marketplaces, in turn questioning the validity of these marketplaces as business models. Image regeneration through iterative prompt refinement is an alternative for users to recreate target images without relying on purchased prompt. %
Additionally, iterative prompt refinement can also enable non-expert users, such as hobbyists or novice designers, to engage and supplement their creative ability, knowledge, and prompt creation skills through experimenting with prompts and image regeneration \cite{tang2024exploring,oppenlaender2024prompting,kulkarni2023word}. %
Beyond these applications, other use cases for image regeneration include digital archiving and art restoration. For example, lost or incomplete visual assets can be recreated from a low-resolution or degraded version of the original image \cite{liang2023iterative}. This technique could be utilized in preserving cultural heritage, or recreating historical imagery where traditional techniques proved difficult. %

Our study builds on these motivations, emphasizing the role of the human in prompt refinement. Prior work on AI-led refinement has explored how automated systems can optimize prompts. For instance, Mañas et al. presents OPT2I, a framework that leverages large language models to automatically refine prompts in \texttt{txt2img} models, improving alignment between prompts and generated images \cite{manas2024improving}. OPT2I iteratively revises user-provided prompts, optimizing a consistency score that evaluates how well the generated image matches the prompt, all without model fine-tuning.
Similarly, a recent study by Zhan et al. on capability-aware prompt reformulation, demonstrate how refining prompt language can significantly enhance image generation quality \cite{zhan2024capability}. By tailoring prompt adjustments to match user proficiency, their system helps users create more coherent and relevant images, regardless of their familiarity with prompt engineering.

In contrast, our study emphasizes the leading role of the human in iterative refinement, investigating how individuals improve their prompts and outputs to achieve desired results. This approach aligns with prior research on human-centric systems, such as \textit{GenAssist} by Huh et al. which provides blind and low-vision users with prompt-guided descriptions and visual verification features to assess generated images’ content and style alignment with initial prompts \cite{huh2023genassist}; but we shift the focus entirely to user-driven adjustments and improvements.

\subsection{Image Similarity Metrics} %

Evaluating image similarity is a fundamental task in computer vision, with applications ranging from image retrieval to quality assessment. Traditional metrics (L2 Euclidean Distance and SSIM) tend to assume pixel-wise independence, and can fall short in capturing the perceptual nuances that come from human judgment. One notable approach that addressed this limitation involves the use of deep features from convolutional neural networks (CNNs). These features are then used to assess image similarity. The Learned Perceptual Image Patch Similarity (LPIPS) metric, or simply Perceptual Similarity (PS), utilized this approach. Specifically, it computes feature embeddings from multiple convolutional layers of pre-trained networks (e.g., AlexNet, VGG, or SqueezeNet), normalizes these activations channel-wise, and then calculates the weighted L2 distance between the feature maps. The final similarity score averages these distances across spatial dimensions and layers. This approach exploits the hierarchical nature of CNNs, allowing the metric
to account for higher-order image structures, context-dependent visual patterns, and other nuances that impact how humans perceive image similarity. As a result, this metric tends to align well with human judgment \cite{zhang2018perceptual}.

Another recent approach is the Contrastive Language-Image Pre-training (CLIP) model, which learns to associate textual and visual information. CLIP is pre-trained on large-scale datasets consisting of image-text pairs, allowing it to generalize well across various domains without needing task-specific fine-tuning. The scores produced measure the similarity between two images based on their embeddings generated by the CLIP model, 
without directly referencing the original text prompts \cite{clip,wang2023exploring}. 

In this study, we employ two different variants of CLIP: B32 and L14.
The B32 variant refers to a specific version with a smaller image representation size. This score is used to measure how similar two images are based on the model's understanding of their features.
Similarly, CLIP L14 is another variant of the same model, but with a larger representation, capturing more detailed image features. The score indicates the degree of similarity between two images, with L14 providing a more precise detection of nuanced differences.

Finally, hashing algorithms offer an alternative approach by condensing images into compact binary representations \cite{imagehash}.
This means that the Hamming distance (the number of differing bits) between these hashes is used to indicate similarity. A smaller distance suggests greater similarity between the images, whereas a larger distance indicates more significant differences \cite{krawertzhash}.

\section{Research Questions \& Hypotheses}
\label{sec:goals}

This study investigates the alignment of computational image similarity metrics (ISMs) with human judgment, and subsequently how iterative prompt refinement affects image regeneration tasks. We aim to understand the relationship between iterative prompt refinement and the tools used to measure its effectiveness. The following research questions guide our investigation:

\noindent\textbf{RQ1 - Human Perception of ISMs: } Do humans generally agree that the selected ISMs are reliable numerical heuristics for evaluating whether two images are perceived as similar or different? 

While Zhang \cite{zhang2018perceptual} shows that their proposed metric, Perceptual Similarity, outperforms previous traditional metrics (SSIM, FSIM, L1 or L2 norm) in aligning with human judgments, 
Sinha and Russell \cite{sinha2011perceptually} demonstrate the limitations of ISMs, cautioning users when interpreting their reliability. 
This raises the question of whether humans consistently perceive these metrics as reliable indicators of similarity. Addressing RQ1 supports our study by assessing how humans perceive ISMs as tools to help facilitate iterative refinement in image regeneration tasks. The hypotheses for RQ1 are then as follows:

\begin{itemize}[noitemsep, topsep=0pt, left=0pt]
    \item \textbf{Hypothesis 1.1:} Human raters exhibit moderate to good agreement with the ISMs as evaluative tools for image similarity. %
    \item \textbf{Hypothesis 1.2:} There are no significant differences across the ISMs in terms of agreement with subjective human assessment.
\end{itemize}

\noindent\textbf{RQ2 - Impact of Iterative Prompt Inference on Image Regeneration:} Given the task of prompt inference to regenerate a target image, does iterative prompt inference improve the ISM score of a user-generated image, meaning it is more similar to the target image?

Building on prior work demonstrating the potential of iterative prompt learning to enhance image alignment with target outputs \cite{liang2023iterative,manas2024improving,zhan2024capability}, we investigate RQ2 in order to determine whether iterative refinement of user-generated prompts results in generated images that more closely match target images as measured by ISMs. The following are hypotheses tied to this research question:

\begin{itemize}[noitemsep, topsep=0pt, left=0pt]
    \item \textbf{Hypothesis 2.1:} Iterative prompt refinement improves the similarity of user-generated images with target images. %
    \item \textbf{Hypothesis 2.2:} Improvement in ISM scores diminishes over successive iterations. %
    \item \textbf{Hypothesis 2.3:} Users perceive images generated from later iterations as more similar with target images compared to earlier iterations.
\end{itemize}

The tests conducted to evaluate these hypotheses are further explained in \Cref{sec:eval}.

\section{Survey Design}
\label{sec:survey}

\subsection{Task: Image Regeneration through Iterative Prompt Refinement}
\label{sec:survey-main-task-1}

Each participant was assigned a set of 10 target images, chosen from the image dataset, and tasked with refining prompts over 10 iterations per image, in order to reproduce the target image. For half of the target images, ISM feedback was displayed to participants (\cref{fig:p0} in \cref{appendix:surveydetails}), while the remaining half omitted this feedback (\cref{fig:p1} in \cref{appendix:surveydetails}) to assess its influence on performance. To ensure the chosen ISMs are tested equally, our participant pool of 20 was divided into 4 subsets of 5, each subset testing a different image similarity metric.

\subsection{Task: Subjective Similarity Ranking}
\label{sec:survey-main-task-2}

Following the iterative refinement task, participants ranked the 10 images generated across iterations for each target image. Rankings were performed using a drag-and-drop interface, with the most similar images placed on the left and the least similar on the right. \cref{fig:p2} in \cref{appendix:surveydetails} shows an example of a ranking page.

\section{Experimental Setup \& Evaluation}
\label{sec:eval}

\subsection{RQ1: Evaluating Alignment}
\label{sec:ism-eval}
Although our primary focus is ultimately on the human assessment of image quality, we begin by examining whether the ISMs can approximate human perceptions of similarity to aid in quantitative analysis. In our dataset, each \textit{target prompt} was rated by a human and by a given ISM. By comparing these sets of ranks, we investigate whether an ISM’s ordering of images correlates meaningfully with how humans order them. If an ISM score aligns with human judgments, it may serve as a useful heuristic in identifying which images are more (or less) similar to the target. However, it is important to approach these results with caution, as no automated metric can fully capture the nuanced ways in which humans evaluate images.

\noindent\textbf{Intraclass Correlation Coefficient:} 
To evaluate alignment of the chosen ISMs with subjective human ratings, we employed the Intraclass Correlation Coefficient (ICC). This metric is specifically designed to assess the degree of agreement or consistency among raters who evaluate the same set of items. Unlike the Pearson correlation, the ICC takes into account not only the linear relationship but also the consistency in how items are scored. In our study, we used a two-way mixed-effects model with a consistency definition, treating the ISM ranking as a fixed effect (since it is a specific algorithm whose performance we want to evaluate) and the images as random effects. This model is suitable because we are interested in whether an ISM's relative ordering of items parallels that of human raters, rather than whether the machine matches human scores exactly.

We interpret the ICC using conventional guidelines for reliability \cite{koo2016guideline}: values below 0.5 are poor, values between 0.5 and 0.75 are moderate, values between 0.75 and 0.90 are good, and values greater than 0.90 are excellent. An unacceptably low ICC value would suggest that an ISM has little resemblance to the human rankings (i.e., the metric is not a good heuristic), whereas a moderately high or better ICC suggests that the ISM reflects a useful, though still imperfect, approximation of human perceptions.

For our analysis, we computed ICC values for different metric types to determine which ones align most closely with human judgments. \Cref{tab:icc} summarizes the ICC results for the four ISMs we evaluated. All metrics exhibit statistically significant ICC values at $p<.001$, indicating that each metric aligns with human ratings at levels significantly above zero agreement; however, their degree of alignment significantly varies. 

\begin{table}[h!]
    \small
    \centering
    \caption{ICC for each ISM. A two-way mixed-effects model with a consistency definition was used, treating items (images) as random effects and each ISM as a fixed effect.}
    \begin{tabular}{lcccc}
        \toprule
        \textbf{ISM} & \textbf{ICC} & \textbf{95\% CI} & \textbf{df} & \textbf{\(p\)-value}\\
        \midrule
        \textbf{PS}   & 0.686 & (0.625, 0.737) & 489 & \(<.001\)\\
        \textbf{B32}  & 0.620 & (0.547, 0.681) & 499 & \(<.001\)\\
        \textbf{L14}  & 0.527 & (0.437, 0.603) & 499 & \(<.001\)\\
        \textbf{ImageHash} & 0.250 & (0.104, 0.372) & 489 & \(<.001\)\\
        \bottomrule
    \end{tabular}
    \label{tab:icc}
\end{table}

We found that B32, L14, and Perceptual Similarity (PS) attained moderate agreement with human raters. Their respective ICC values exceed 0.50, suggesting that while they may not perfectly replicate human judgments, they sufficiently capture a meaningful portion of how humans perceive image similarity for the purpose of this study. Consequently, these metrics can reasonably serve as proxies for human preferences in subsequent analyses.

In contrast, ImageHash yielded a notably lower ICC of 0.250, signifying poor alignment with human evaluations. Because our aim is to ensure that each ISM used in the study reflects human judgments to an acceptable degree, we have decided to exclude ImageHash from the next stage of analysis. By removing ImageHash, we focus our analyses on those metrics that offer more credible approximations of human-driven perceptions of image similarity.

\begin{leftbar} \noindent
The ICC values in \cref{tab:icc} demonstrate that \textbf{B32}, \textbf{L14}, and \textbf{PS} achieve moderate alignment with human judgments, which supports \textbf{Hypothesis~1.1}.
However, the notably lower ICC for \textbf{ImageHash} (0.250) suggests that \emph{not all} metrics yield equivalent alignment. Thus, \textbf{Hypothesis~1.2}, predicting no significant differences among the ISMs, holds only for B32, L14, and PS, but not for ImageHash. This is why ImageHash was excluded from further modeling, ensuring we focus subsequent analyses on the ISMs that fulfill Hypothesis~1.1’s criterion of moderately capturing human judgments.
\end{leftbar}

\subsection{RQ2: Evaluating Refinement}
\label{sec:iterative-eval}

\noindent\textbf{Mixed-Effects Model Results:}
After excluding the ImageHash metric, we fit a linear mixed‐effects model, defined in \Cref{sec:model-specs} to examine how our fixed factors and random intercepts contribute to the adjusted ISM. Table~\ref{tab:fixed-effects} presents the Type~III Tests of Fixed Effects. We observe that:

\begin{table}[b]
    \centering \small
    \setlength{\tabcolsep}{3pt} %
    \caption{Type III Tests of Fixed Effects for the mixed-effects model predicting adjusted scores.}
    \begin{tabular}{lrrrr}
        \toprule
        \textbf{Effect}        & \textbf{Num df} & \textbf{Den df} & \textbf{F}      & \textbf{Sig.} \\
        \midrule
        Intercept             & 0               & --              & --              & --            \\
        iteration            & 9               & 1451            & 11.486          & $<.001$*       \\
        gender               & 1               & 1451            & 0.001           & .999          \\
        education            & 2               & 1451            & 0.001           & .999          \\
        native language      & 1               & 1451            & 0.253           & .615          \\
        \texttt{txt2img} familiarity & 1        & 1451            & 0.253           & .615          \\
        subject              & 4               & 1451            & 5.758           & $<.001$*       \\
        visibility of metric         & 1               & 1451            & 2.061           & .151          \\
        iteration$\times$visibility of metric & 9      & 1451            & 0.515           & .865          \\
        type of metric          & 2               & 1451            & 0.446           & .504          \\
        \bottomrule
    \end{tabular}
    \label{tab:fixed-effects}
\end{table}

\begin{itemize}[noitemsep, topsep=0pt, left=0pt]
    \item \textbf{Iteration} is significant. We observe $(F(9,1451)=11.486,p<.001)$, suggesting that the adjusted score changes systematically across successive iterations. Post-hoc comparisons of the iteration estimates (\cref{tab:fixed-effects}) indicate an overall trend toward improved adjusted scores over the first several iterations, with diminishing effects after iteration 6. From \cref{tab:iteration-fixedeffects}, we see that iterations 1 through 6 each exhibit significantly lower (improved) adjusted scores compared to the reference level (iteration 10). Specifically, iteration 1 has the largest negative estimate (-0.053), and although the magnitude of improvement tapers with increasing iteration number, all estimates remain significantly different from zero through iteration 6. By iteration 7 and beyond, the effect is no longer statistically significant, implying that user performance begins to plateau around the seventh iteration.
    \item \textbf{Subject}, the content of the prompt, (e.g., ``cat,'' ``astronaut,'' etc.) also shows a significant main effect $(F(4,1451)=5.758,p<.001)$. This indicates that some subjects inherently tend to yield higher or lower adjusted scores, irrespective of other predictors.
    \item \textbf{Visibility of ISM} is not significant $(F(1,1451)=2.061,p=.151)$, nor is the interaction between \textbf{iteration} and \textbf{visibility of metric} $(F(9,1451)=0.515,p=.865)$. Thus, showing the metric during the task does not reliably alter the rate of improvement across iterations in this dataset.
    \item \textbf{The specific type of ISM} (PS, CLIP L14, or CLIP B32) shown to a user does not exhibit a statistically significant main effect under this model $(F(2,1451)=0.446,p=.504)$. Given that we already established all three to have acceptable alignment with human rankings via the ICC, this result suggests that once the model controls for other factors, the three ISMs produce broadly similar adjusted scores on average.
\end{itemize}

\Cref{tab:cov-params} shows the estimates of the random effects and the AR(1) correlation in the residuals. We included random intercepts for session\_id and target\_prompt to account for unexplained variability at both the participant and prompt level. Both variance components are small but highly significant ($p<.001$), indicating that individual differences between sessions and systematic differences between prompts do exist. Additionally, the AR(1) correlation $\rho=-0.230$ is statistically significant ($p<.001$). Although negative serial correlation may seem counterintuitive, it indicates that if a participant’s adjusted score is above the model’s prediction at one iteration, it tends to be slightly below the model’s prediction at the next iteration (and vice versa).

Overall, these results suggest that \textbf{iteration} and \textbf{subject} have robust influences on adjusted scores, whereas the \textbf{visibility of metric} and the \textbf{type of metric} do not produce strong differential effects. The random‐effects estimates affirm that allowing each participant and each target prompt to vary with its own intercept meaningfully improves model fit.
These findings inform our subsequent interpretations of user performance. Because iteration consistently emerges as a key predictor, our data suggest that participants’ scores improve over time. Meanwhile, the negative AR(1) coefficient indicates small but significant oscillations from iteration to iteration in how users respond. 

\begin{table}[h!]
    \centering \small
    \setlength{\tabcolsep}{3pt} %
    \caption{Selected parameter estimates for significant fixed effects in the mixed-effects model. The reference category for iteration is iteration = 10.}
    \begin{tabular}{lrrrr}
    \toprule
    \textbf{Effect} & \textbf{Estimate} & \textbf{Std. Error} & \textbf{t} & \textbf{p-value}\\
    \midrule
    \multicolumn{5}{l}{\textbf{Fixed Effects (Iteration)}} \\
        \quad Intercept     & 0.620    & 0.080    & 7.75   & $<.001$*\\
        \quad iteration = 1 & $-0.053$ & 0.010    & $-5.28$& $<.001$*\\
        \quad iteration = 2 & $-0.046$ & 0.010    & $-4.45$& $<.001$*\\
        \quad iteration = 3 & $-0.032$ & 0.010    & $-3.18$& .001*   \\
        \quad iteration = 4 & $-0.025$ & 0.010    & $-2.51$& .012*   \\
        \quad iteration = 5 & $-0.025$ & 0.010    & $-2.46$& .014*   \\
        \quad iteration = 6 & $-0.023$ & 0.010    & $-2.28$& .023*   \\
        \quad iteration = 7 & $-0.018$ & 0.010    & $-1.80$& .071   \\
        \quad iteration = 8 & $-0.007$ & 0.010    & $-0.73$& .466   \\
        \quad iteration = 9 & $-0.001$ & 0.010    & $-0.09$& .928   \\
        \quad iteration = 10 (ref) & 0 & -- & -- & -- \\
    \bottomrule
    \end{tabular}
    \label{tab:iteration-fixedeffects}
\end{table}

\begin{table}[h]
    \centering \small
    \setlength{\tabcolsep}{4pt}
    \caption{Estimates of covariance parameters for the random effects and AR(1) residual structure.}
    \begin{tabular}{lrrr}
        \toprule
        \textbf{Parameter}         & \textbf{Estimate} & \textbf{Std. Error} & \textbf{Sig.} \\
        \midrule
        AR1 Diagonal  & 0.004 & 0.000 & -- \\
        AR1 $\rho$                          & -0.230 & 0.022 & $<.001$* \\
        Intercept (Variance)             & 0.002 & 0.000 & -- \\
        session\_id (Variance)           & 0.002 & 4.886E-7 & $<.001$* \\
        target\_prompt (Variance)        & 0.005 & 0.000 & -- \\
        \bottomrule
    \end{tabular}
    \label{tab:cov-params}
\end{table}

\begin{leftbar} \noindent
The mixed-effects model results highlight that \textbf{Iteration} emerged as a significant predictor of the adjusted ISM score, with iterations~1 through~6 each showing improved scores relative to iteration~10. This directly supports \textbf{Hypothesis~2.1}: successive iterations lead to meaningful gains in similarity with the target image. Moreover, these improvements diminish beyond iteration~6, suggesting a plateau effect consistent with \textbf{Hypothesis~2.2}. Taken together, these patterns indicate that most of the iterative benefit is realized in the earlier cycles of prompt refinement, after which further iterations yield less significant enhancements.
\end{leftbar}

\noindent\textbf{Top User-Ranked Images:}
While the mixed-effects model in the \Cref{sec:model-specs} relies on ISM-based scores, we also examined a purely human-centric measure of iterative improvement. Specifically, from the ranking procedure explained in the \Cref{sec:survey-main-task-2}, we evaluated from which iteration came a user's top-ranked image. If iterative prompting has no effect on the user's perception of which generated image is closest to the target image, we would expect that a user's top-ranked image is equally likely to come from any of the 10 iterations. 

To test this, we aggregated the iteration at which each user selected their top-ranked image into one distribution and performed a chi-square goodness-of-fit test against the null hypothesis of a uniform distribution. \Cref{tab:top-chisq} shows the observed and expected frequencies per iteration. The result was highly significant, $\chi^2(9) = 71.200, p < .001$, indicating that users most frequently selected images generated in later iterations (particularly 9 and~10), rather than early ones. In other words, the best image rarely appeared in the initial iterations, providing additional evidence that users iteratively refine their prompts over time to achieve better results.

\begin{table}[htb]
    \small
    \centering
    \caption{Chi-square test of the iteration chosen as a user's top-ranked image. Under the null hypothesis of a uniform choice across 10 iterations (Expected~N~=~15 each), the last two iterations (9 and 10) were chosen far more often than expected.}
    \begin{tabular}{lrrr}
        \toprule
        \textbf{Iteration} & \textbf{Observed N} & \textbf{Expected N} & \textbf{Residual} \\
        \midrule
        1  & 12 & 15 & -3  \\
        2  & 9  & 15 & -6  \\
        3  & 9  & 15 & -6  \\
        4  & 7  & 15 & -8  \\
        5  & 10 & 15 & -5  \\
        6  & 11 & 15 & -4  \\
        7  & 13 & 15 & -2  \\
        8  & 14 & 15 & -1  \\
        9*  & 21 & 15 & 6*   \\
        10* & 44 & 15 & 29*  \\
        \midrule
        \textbf{Total} & 150 & 150 & -- \\
        \midrule
        \textbf{Asymp. Sig.} & & & $p<0.001$ \\
        \bottomrule
    \end{tabular}
    \label{tab:top-chisq}
\end{table}

\begin{leftbar} \noindent
This human-only perspective, independent of ISM scores, provides evidence and support for Hypothesis 2.3, that iterative prompting contributes to users' subjective sense of improvement in the image regeneration task. Users predominantly favored images from their final two iterations. 
\end{leftbar}

\section{Discussion}
\label{sec:discussion}

\noindent\textbf{Implications from Results:}
Generative AI has rapidly expanded, making prompt engineering an increasingly critical skill \cite{oppenlaender2024prompting}. Our findings confirm that single-shot prompting often falls short when aiming for a precise target visual; in contrast, iterative prompt refinement offers a more positive path to alignment.
From the results of our analysis in \Cref{sec:eval}; the following insights stand out:

\begin{itemize}[noitemsep, topsep=0pt, left=0pt]
    \item \textbf{Alignment between subjective human assessment and objective ISM scores:} Moderate agreement between human assessment and ISMs, as demonstrated by the ICC values in \Cref{sec:ism-eval}, demonstrates that ISMs like Perceptual Similarity (PS) and CLIP variants can reasonably approximate human perception. This is further supported by research such as Ghildyal and Liu, which demonstrated a new metric based on PS that is robust to small misalignments in aligning with human perception \cite{ghildyal2022shift}. Zhang and Krawetz have shown that that pixel-wise image comparison tend to not align well with human perception, which correlates to ImageHash's ICC being the lowest in our results \cite{zhang2018perceptual,krawertzhash}. Overall, this has potentials for the development of educational tools, where these metrics could guide novice users in refining prompts to achieve a desired AI-generated images. %
    
    \item \textbf{Effectiveness of iterative prompt refinement:} The results from the mixed-effects model in \Cref{sec:iterative-eval} serves as an objective demonstration that iterative refinement meaningfully improves the alignment of user-generated images with target images, as evidenced by the improvement in ISM over the first six iterations, followed by a plateau in score until the last iteration. This plateau could suggest that participants show the most improvement in the earlier iterations, with a potential for a diminishing return in the later iterations. Moreover, users predominantly favor images from their final two iterations, meaning there is a subjective sense of improvement in their image regeneration task. Overall, the results confirms the utility of iterative approaches of human when performing generative AI-related tasks; but also highlight a potential for further research on enhanced guidance or feedback support for users to overcome the improvement plateau.
    Since previous research has shown effectiveness in AI-led prompt refinement \cite{manas2024improving,zhan2024capability,liang2023iterative}; our results can potentially open up avenues in research for human-AI collaboration in iterative prompt refinement. %
\end{itemize}

\noindent\textbf{Limitations and Future Work:}\quad Despite promising initial evidence, our study faces several constraints. First, the participant pool of 20 university students was small and homogeneous; future research could address this by recruiting larger and more diverse samples to uncover population-specific insights. 
Second, participants were only given a fixed number of iterations per target image, which may not reflect the broader potential of iterative refinement and may have have introduced potential biases in the later iterations, as suggested by by trends in the top user-ranked images (\Cref{sec:iterative-eval}). Subsequent work could adopt flexible iteration counts or investigate the optimal point at which performance gains taper off. 
Third, while we controlled for the influence of different \texttt{target\_prompt} conditions, assigning the same set of prompts to all participants could help clarify whether certain topics inherently yield lower or higher ISM scores (\Cref{sec:iterative-eval}).We also observed that simply providing ISM feedback had little effect on outcomes, indicating a need for more intuitive and user-friendly feedback mechanisms that better guide users toward improved results.

Beyond these considerations, our focus on \texttt{txt2img} models leaves open the question of whether the benefits of iterative prompt refinement extend to other generative domains such as large language models, code generation, and audio synthesis. Research in these areas would help verify the broader applicability of our findings. Finally, although improving a user’s ability to regenerate visuals through iterative refinement carries many creative opportunities, it also raises ethical and societal concerns related to misinformation, deepfakes, and intellectual property rights \cite{oecdGenerativeRisks,Marcus_Southen_2024}. Future investigations should examine how best to moderate these risks—whether through training and policy interventions or through human-AI collaboration tools—so that enhanced generative capabilities remain both innovative and responsible.

\section*{Conclusion}
\label{sec:conclusion}
Our study demonstrated that iterative prompt refinement substantially enhances alignment between AI-generated images and target visuals, particularly in the early stages of refinement. 
Moderate agreement between select ISMs and human evaluations indicated that such metrics hold promise as objective feedback tools for user workflows. Nevertheless, limitations related to participant diversity, iteration count, and feedback mechanisms point to the need for further research.
Given the increasing prevalence of AI-generated content on social media platforms as well as the web as a whole, this work provides a solid understanding for optimizing human-centric workflows in generative AI tasks, noting the importance of iterative refinement for achieving desired results.

\bibliographystyle{named}
\bibliography{references}

\appendix

\section{Chosen AI-Image Generation Model}
\label{sec:chosen-model}

Both the target and user images are generated using the Stable Diffusion 3.0 model\footnote{\url{https://huggingface.co/stabilityai/stable-diffusion-3-medium}} from Stability AI. We chose this model it was the most current model when we began experimental design and setup,
and has demonstrated improved performance in areas of visual aesthetics, prompt following, and typography \cite{esserstable}. Additionally, this model offers easier control over the generation parameters compared to other models like DALL-E or Midjourney. Moreover, the model has a Community License\footnote{\url{https://stability.ai/license}} which is appropriate for our research usage.
\cref{tab:sd3-params} lists the parameters that were used in the generation of both target and user images, any parameters not listed were kept at the default value. These chosen parameters offered a balance in image quality and inference time for usage on an Nvidia RTX 4060 Ti.

\begin{table}[h]
    \centering
    \setlength{\tabcolsep}{3pt} %
    \caption{Parameters used for generating images with Stable Diffusion 3.0.}
    \begin{tabular}{lr}
        \toprule
        \textbf{Parameter}         & \textbf{Value} \\ \midrule
        Seed                       & 1029           \\
        Batch Size                 & 1              \\
        Steps                      & 10             \\
        Sampler Index              & Euler          \\
        CFG Scale                  & 5              \\
        Width                      & 1024 px        \\
        Height                     & 1024 px        \\
        Denoising Strength         & 0.4            \\ \bottomrule
    \end{tabular}
    \label{tab:sd3-params}
\end{table}

\section{Prompt and Image Dataset}
\label{sec:dataset}

The images presented to the participants were generated using a set of prompts constructed based on the method employed by Trinh \cite{trinh2024promptly} to create their ``Controlled" dataset. Using this method allowed for consistency in variations across prompts, minimizing any potential biases that target prompts may introduce in our participants' performance.
In total, there are 100 prompts, evenly divided across 5 fixed subjects (\textit{astronaut}, \textit{cat}, \textit{man}, \textit{robot}, and \textit{woman}). These prompts were then used to generate 100 target images for the iterative prompt refinement task of our survey.

\section{Mixed-Effects Model Specification}
\label{sec:model-specs}

To adequately capture both the repeated measurements over time and the hierarchical, nested nature of our data, we specified a mixed-effects model that incorporates fixed effects of interest (e.g., iteration, demographics, familiarity with \texttt{txt2img} models, etc.) and random intercepts for each participant, represented by their unique session id (\(\texttt{session\_id}\)) and the original target prompt (\(\texttt{target\_prompt}\)) \cite{moerbeek2004consequence}. For clarity, each human participant was identified via a unique \(\texttt{session\_id}\), and each participant was given multiple \(\texttt{target\_prompts}\) over the course of the study. For each \(\texttt{target\_prompt}\), 10 repeated iterations were collected, thereby yielding multiple observations within each \(\texttt{session\_id}\). This design inherently involves both repeated measures (multiple iterations within a subject) and hierarchical structure (observations nested within target prompts, which are in turn nested within participants).

To account for variability at the subject level, we included a random intercept for \(\texttt{session\_id}\), thereby acknowledging that some subjects may systematically differ in their responses regardless of other predictors. Similarly, a random intercept for \(\texttt{target\_prompt}\) controls for potential variation attributable to the unique characteristics of each prompt. Together, these random effects allow us to partition out unexplained variability that arises from both human-level and prompt-level differences, improving the model’s accuracy and interpretability of the fixed effects.

Given that each human subject encountered multiple \(\texttt{target\_prompts}\) sequentially, we anticipated temporal autocorrelation in the residuals (e.g., consecutive iterations may be more similar to each other than to those taken far apart in time). To address this, we imposed an autoregressive, AR(1) covariance structure on the residuals. The AR(1) specification assumes that errors from measurements that are close together in time are more strongly correlated than errors from measurements that are further apart, which more accurately reflects the within-subject temporal dynamics of our repeated-measures design. Taken together, these choices (mixed-effects specification, nested and repeated factors, and AR(1) covariance) yield a model that properly accounts for both fixed and random sources of variation in the data, ensuring that inferences drawn from the fixed-effect parameters remain unbiased. The equation below represents how the model measured the adjusted ISM score, based on various coefficients that account for fixed effects, random effects, and autoregressive correlations between iterations.

\begin{align*}
y_{ijkm} 
&= \beta_0 
   + \sum_{r=0}^{9} \beta_r \, \text{Iter}_r 
   + \sum_{d=0}^{D} \beta_{1d} \, \text{Demo}_{id} \\
&\quad + \sum_{u=0}^{3} \beta_{2u} \, \text{Fam}_u 
   + \sum_{v=0}^{4} \beta_{3v} \, \text{Sub}_v 
   + \beta_4 \, \text{Met}_i \\
&\quad + \sum_{r=0}^{9} \beta_{5r} \, (\text{Iter}_r \times \text{Met}_i)
   + \beta_6 \, \text{Typ}_i 
   + b_j 
   + b_k \\
&\quad + \varepsilon_{ijkm}
\end{align*}

\noindent
\textbf{Definitions of Coefficients:}
\begin{itemize}
    \item $y_{ijkm}$: Adjusted\footnote{to a range of 0 to 1, with higher score meaning better similarity} ISM score measured for the $i$-th participant, in the $j$-th session, for the $k$-th target prompt, at the $m$-th iteration
    \item \(\beta_0\): Global Intercept
    \item \(\beta_r\): Fixed effect of Iteration (\(r = 0,\dots,9\))
    \item \(\beta_{1d}\): Fixed effects for Demographics (\(d = 0, \dots, D\), where \(D\) includes variables like Gender, College Major, and if English is native language)
    \item \(\beta_{2u}\): Fixed effect for \texttt{txt2img} Familiarity (\(u = 0,\dots,3\))
    \item \(\beta_{3v}\): Fixed effect of Subject (\(v = 0,\dots,4\))
    \item \(\beta_4\): Fixed effect of Visibility of Metric (binary)
    \item \(\beta_{5r}\): Interaction effect of Iteration \(\times\) Visibility of Metric (\(r = 0,\dots,9\))
    \item \(\beta_6\): Fixed effect of Metric Type (categorical)
\end{itemize}

\noindent
\textbf{Random Effects:}
\begin{itemize}
    \item \(b_j\): Random intercept for \(\texttt{session\_id}\), capturing variability across sessions, with \(b_j \sim N(0, \sigma^2_{b,j})\)
    \item \(b_k\): Random intercept for \(\texttt{target\_prompt}\), capturing variability across prompts, with \(b_k \sim N(0, \sigma^2_{b,k})\)
\end{itemize}

\noindent
\textbf{Repeated Measures:}
\begin{itemize}
    \item Repeated measurements are made at each combination of \(\text{Iteration}\) and \(\texttt{target\_prompt}\) within each \(\texttt{session\_id}\).
    \item Residuals \(\varepsilon_{ijkm}\) follow an AR(1) covariance structure to account for the temporal correlation between repeated measures:
    \[
    \text{Cov}(\varepsilon_{ijkm}, \varepsilon_{ijk'm'}) = 
    \begin{cases}
    \sigma^2_{e}, & \text{if } (m = m'),\\
    \sigma^2_{e}\,\rho^{|m - m'|}, & \text{if } (m \neq m'),
    \end{cases}
    \]
    where \(\rho\) is the AR(1) correlation parameter, which governs how strongly correlated residuals are across time. The exponent ${|m - m'|}$ is how many iterations apart two observations are. If the measurements are two steps apart, the correlation becomes $\rho^2$, and so on. This exponential decay captures the intuition that as measurements get further apart, their correlation decreases. $\sigma^2_{e}$ is the residual variance, or the variance of the model error term $\varepsilon$.
\end{itemize}

\noindent
\textbf{Estimation:} Restricted maximum likelihood (REML).

\section{Demographic and Familiarity with Generative AI}
\label{sec:presurvey-demographic}

\begin{figure}[H]
\centering
\begin{subfigure}[b]{0.49\linewidth}
    \centering
    \includegraphics[width=\textwidth]{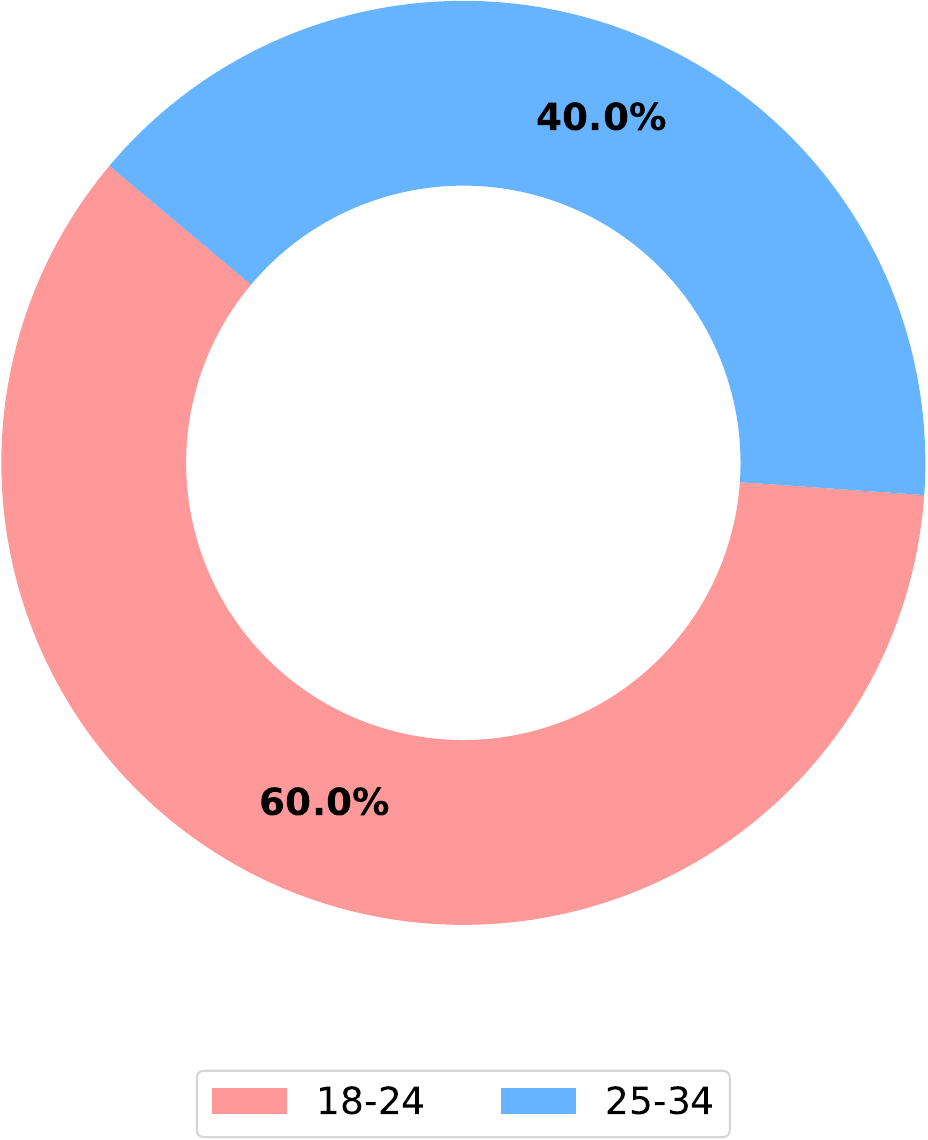}
    \caption{Age distribution.}
    \label{fig:age-distribution}
\end{subfigure}
\hfill
\begin{subfigure}[b]{0.49\linewidth}
    \centering
    \includegraphics[width=\textwidth]{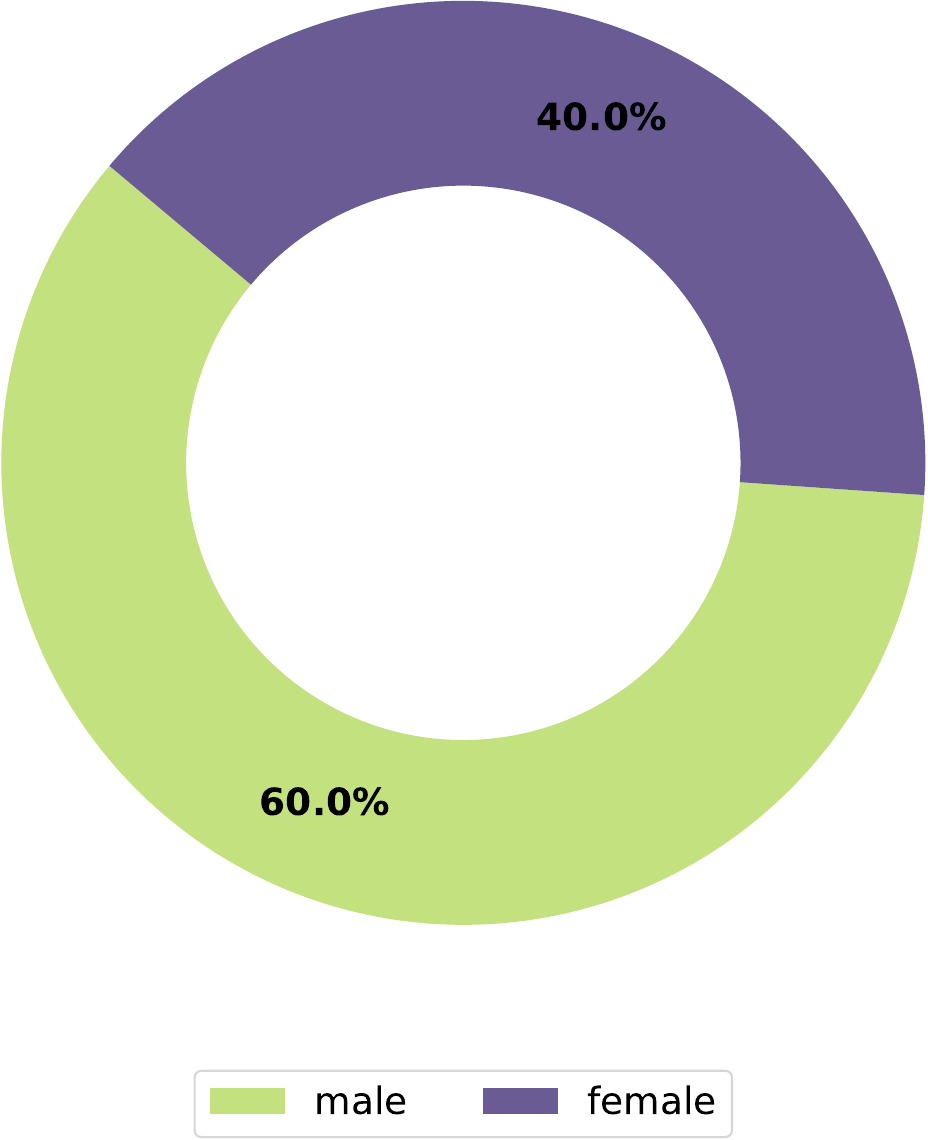}
    \caption{Gender distribution.}
    \label{fig:gender-distribution}
\end{subfigure}
\caption{Demographic distribution of survey participants.}
\label{fig:demographics}
\end{figure}

Participants were all affiliated with our host institution. The majority were aged between 18 - 24 years (65\%), while the remaining 35\% fell in the 25 - 34 age range. Gender distribution was relatively balanced, with slightly over half of our participants identifying as male (55\%), while the remaining identified as female (45\%). \cref{fig:demographics} shows this distribution.

Most (85\%) participants reported having English as their native language.
Regarding academic background, the majority of our participants (40\%) were from ``\emph{Computer Science}'', while the rest are participants in fields like ``\emph{Data Science}'', ``\emph{Electrical Engineering}'', ``\emph{Art}'', etc.

\begin{figure}[H]
\centering
\includegraphics[width=\linewidth]{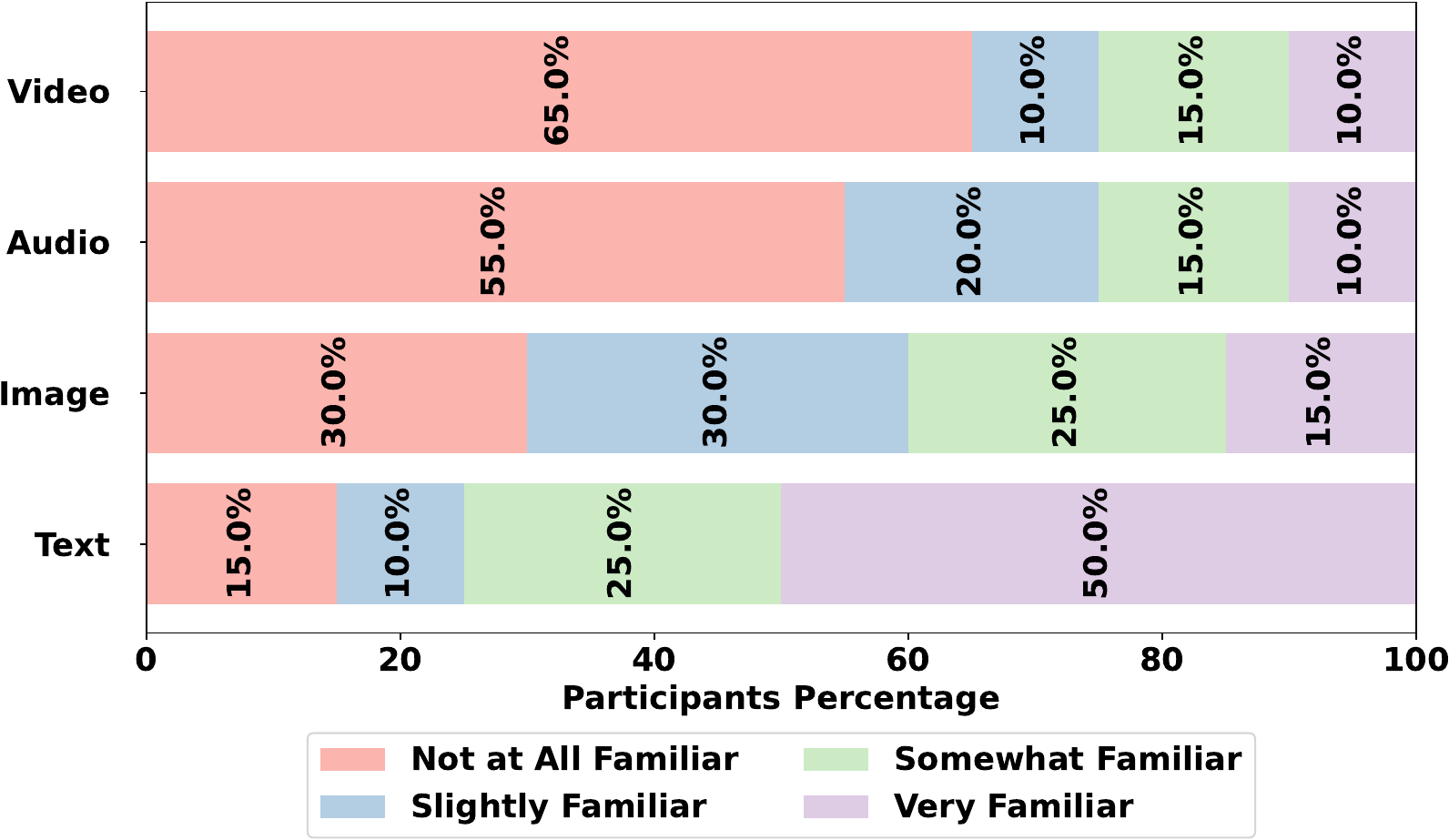}
\caption{Participants' familiarity levels with different generative AI tools.} %
\label{fig:ai-tool-familiarity}
\end{figure}

\Cref{fig:ai-tool-familiarity} shows the familiarity  levels of the participants across media types, with text generative tools being the most familiar to participants, as 50\% reported that they are ``\emph{Very Familiar}''. Image generation tools follow with a more balanced distribution, where most participants are ``\emph{Slightly}'' (30\%) or ``\emph{Somewhat Familiar}'' (25\%). In contrast, familiarity with audio and video generative tools is significantly lower, with over half of the participant pool being ``\emph{Not at All Familiar}'' with these mediums, while only 10\% reported being ``\emph{Very Familiar}''.  These results suggest that text and image generative models are more accessible or commonly encountered, while audio and video generative models may require additional exposure or expertise. Overall, collecting participants' familiarity with generative AI tools may reveal potential variability in performance of their iterative prompt refinement.

\section{Priming}
\label{sec:priming}

Before partaking in the main iterative prompt refinement task of the survey, participants were primed on which ISM they would encounter as participants were randomly assigned one of the four ISMs as part of the experimental design. This priming included explanations of how the metrics work, their value ranges, and whether higher or lower values indicate a closer similarity. This understanding is tested by a short quiz of multiple choice questions on the value of the metric between two images. The participant must pass the quiz in order to be eligible to continue the survey. Priming aligns with prior research suggesting that targeted instruction improves cognitive alignment with task goals \cite{chanpriming}, thereby enabling participants to potentially use ISMs more effectively during refinement.

\section{Data Collection}
\label{appendix:data-collection}

\subsection{Participant Recruitment and Data Handling}
\label{appendix:consent}

The study received an \textit{Exempt} status from our host institution’s IRB, as it posed minimal risks to participants. All participants were required to sign a consent form confirming they were over 18 years old and aware of any potential risks associated with the survey.

Survey responses, including demographic data, were collected anonymously. Unique session IDs were used to correlate deidentified survey answers with demographic information. The only personally identifiable information collected — participant names and email addresses — was used solely for distributing gift cards and was not tied to participants' responses.

All collected data was securely stored on our lab's server. Access to this data was restricted to researchers directly involved in the project to maintain confidentiality and security.

\subsection{Survey Responses Statistics}
\label{sec:survey-stats}

The survey was conducted from late October to mid-December 2024, with 20 participants completing the tasks. Overall, the study recorded 2000 prompts and their associated similarity metric scores, alongside subjective rankings for all participant-generated images. Data were securely stored and de-identified, and participants received \$20 gift cards as compensation for, on average, an hour of their time. 

\subsection{Survey Instructions and Screenshots} 
\label{appendix:surveydetails}

\begin{figure*}[h]
  \centering
  \includegraphics[width=\linewidth]{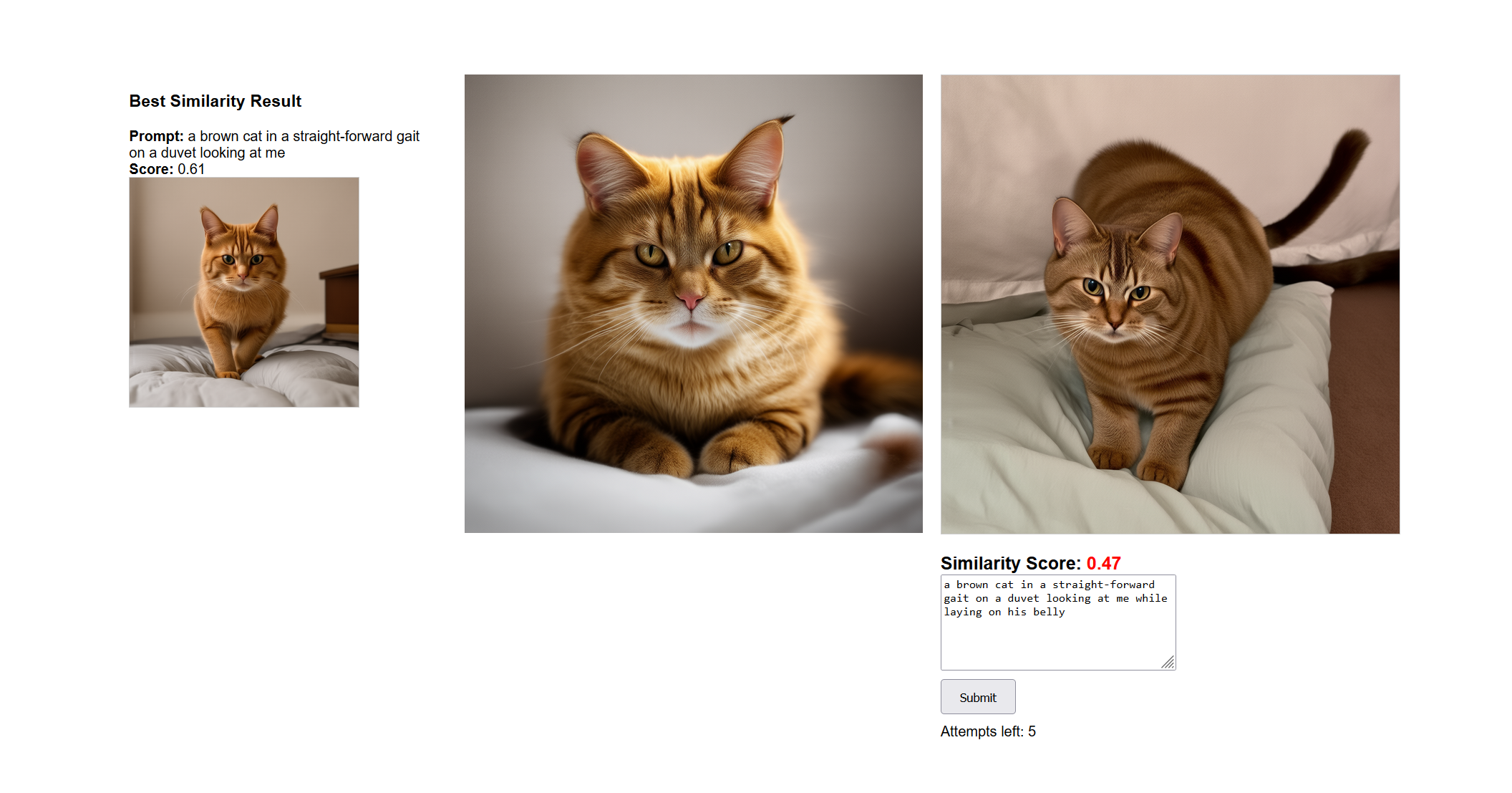}
  \caption{An example of the iterative prompt refinement task with the ISM score feedback. The participant submits their prompt in the box, they will then see the generated image and the associated similarity score above. A green similarity score means the current iteration is better than the last, while a red similarity score means the opposite. To the left of the target image is an area to keep track of their best effort, including their best prompt and best similarity score.}
  \label{fig:p0}
\end{figure*}

\begin{figure*}[h]
  \centering
  \includegraphics[width=0.8\linewidth]{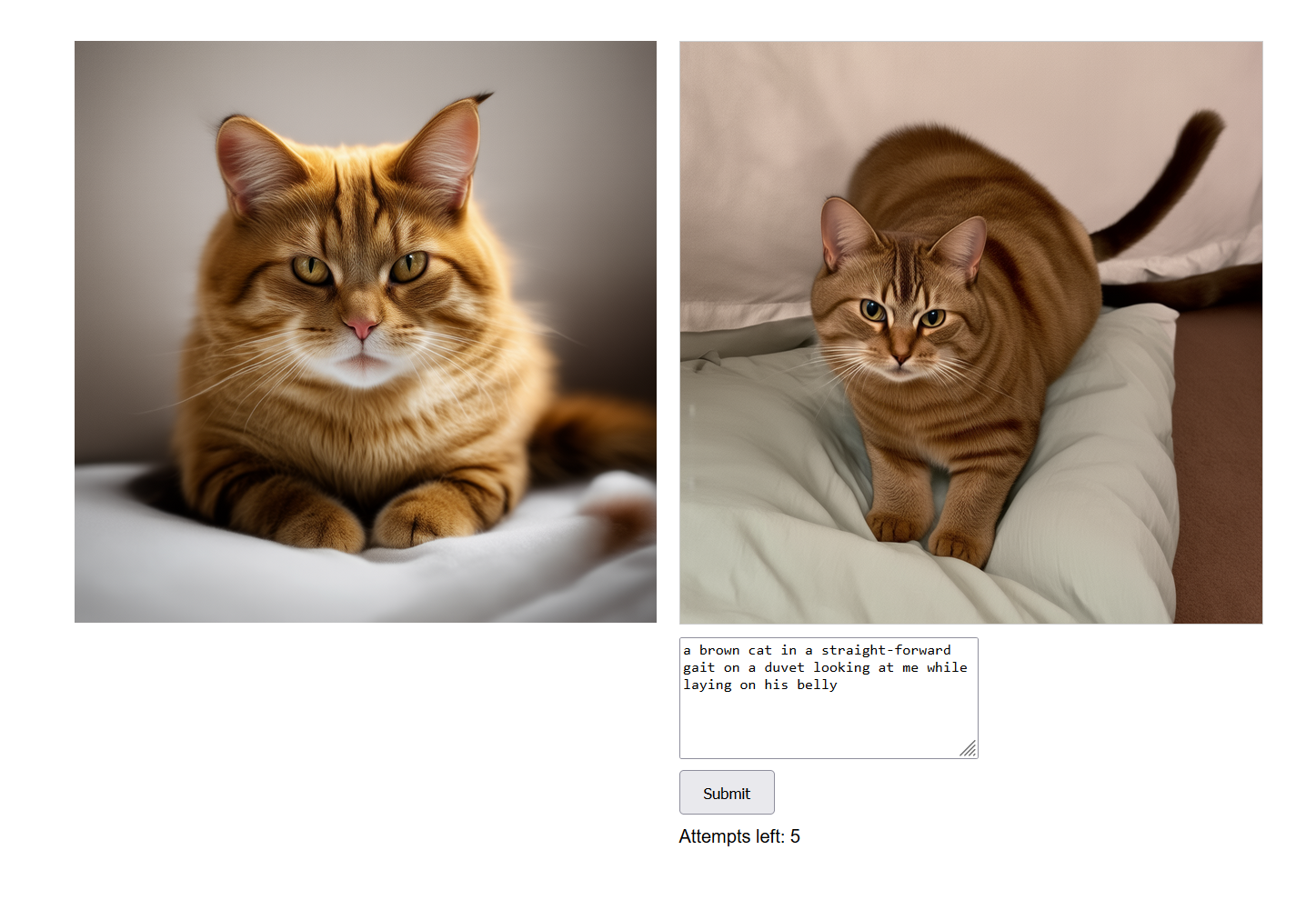}
  \caption{An example of the iterative prompt refinement task without the ISM score feedback. The participant submits their prompt in the box, they will then see the generated image above.}
  \label{fig:p1}
\end{figure*}

\begin{figure*}[h]
  \includegraphics[width=\linewidth]{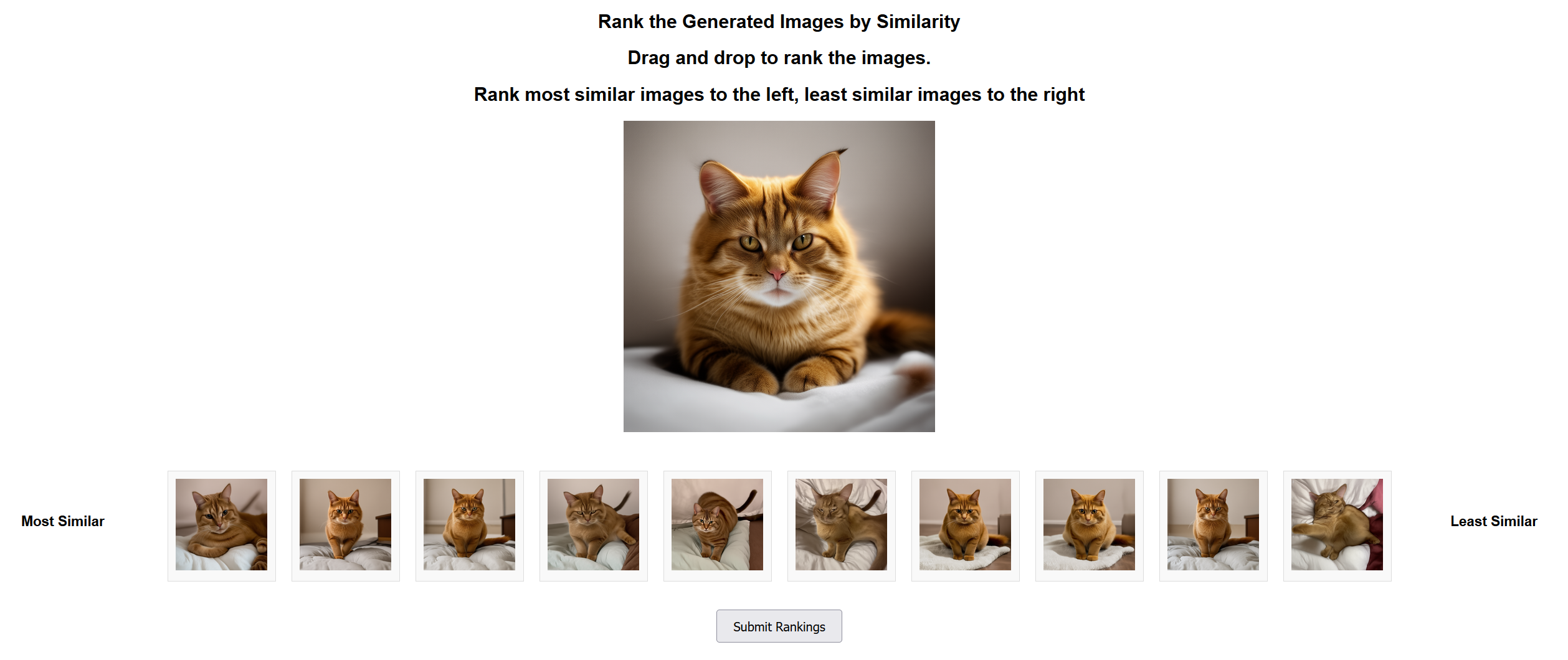}
  \caption{An example of the ranking task. The participant must drag and drop the 10 generated images in order of Most Similar to the left, and Least Similar to the right. The target image is above the set of generated images.}
  \label{fig:p2}
\end{figure*}

\Cref{fig:p0,fig:p1,fig:p2} in this section showcase the main parts of the survey interface for image regeneration, featuring examples of the iterative prompt refinement task both with and without ISM feedback, along with the image ranking task.

\label{appendix:ranking-task}

\end{document}